\documentclass[11pt,letterpaper]{article}
\usepackage{emnlp2017}
\usepackage{times}
\usepackage{latexsym}
\usepackage{url}
\usepackage{xcolor}
\usepackage{colortbl}
\usepackage{amsmath}
\usepackage{tikz}
\usetikzlibrary{arrows,chains,decorations.pathreplacing,fit,matrix,positioning}

\emnlpfinalcopy

\def\emnlppaperid{***}

\newcommand{\example}[1]{\textit{#1}}
\newcommand{\newterm}[1]{\textbf{#1}}
\newcommand{\shapeworld}{\textsc{ShapeWorld}}
\newcommand{\githuburl}{\url{https://github.com/AlexKuhnle/ShapeWorld}}

\newcommand{\muchbetter}{\cellcolor{green!55}}
\newcommand{\better}{\cellcolor{green!35}}
\newcommand{\worse}{\cellcolor{red!35}}
\newcommand{\muchworse}{\cellcolor{red!55}}
\newcommand{\inconsistent}{\cellcolor{blue!25}}

\title{\shapeworld: A new test methodology for multimodal language understanding}

\author{Alexander Kuhnle \\
University of Cambridge \\
{\tt aok25@cam.ac.uk} \\\And
Ann Copestake \\
University of Cambridge \\
{\tt aac10@cam.ac.uk} \\}

\begin{document}
\maketitle
\begin{abstract}
We introduce a novel framework for evaluating multimodal deep learning models with respect to their language understanding and generalization abilities. In this approach, artificial data is automatically generated according to the experimenter's specifications. The content of the data, both during training and evaluation, can be controlled in detail, which enables tasks to be created that require \emph{true} generalization abilities, in particular the combination of previously introduced concepts in novel ways. We demonstrate the potential of our methodology by evaluating various visual question answering models on four different tasks, and show how our framework gives us detailed insights into their capabilities and limitations. By open-sourcing our framework, we hope to stimulate progress in the field of multimodal language understanding.
\end{abstract}

\section{Introduction}

\begin{figure*}
\begin{center}
\footnotesize
\begin{minipage}{0.11\linewidth}
\centering
\phantom{training}\par
\includegraphics[width=\linewidth]{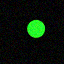}\par
\scriptsize{\example{There is a green circle.}}\par
\textcolor{green}{True}\par\ 
\end{minipage}\,\begin{minipage}{0.11\linewidth}
\centering
training\par
\includegraphics[width=\linewidth]{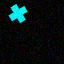}\par
\scriptsize{\example{There is a green cross.}}\par
\textcolor{red}{False}\par\ 
\end{minipage}\,\begin{minipage}{0.11\linewidth}
\centering
\phantom{training}\par
\includegraphics[width=\linewidth]{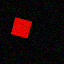}\par
\scriptsize{\example{There is a red square.}}\par
\textcolor{green}{True}\par\ 
\end{minipage}$\,\Rightarrow\,$\begin{minipage}{0.11\linewidth}
\centering
evaluation\phantom{g}\par
\includegraphics[width=\linewidth]{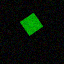}\par
\scriptsize{\example{There is a green square.}}\par
\textcolor{orange}{???}\par\ 
\end{minipage}
\hspace{0.3cm}
\begin{minipage}{0.11\linewidth}
\centering
\phantom{training}\par
\includegraphics[width=\linewidth]{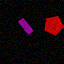}\par
\scriptsize{\example{An ellipse is to the left of a red pentagon.}}\par
\textcolor{red}{False}
\end{minipage}\,\begin{minipage}{0.11\linewidth}
\centering
training\par
\includegraphics[width=\linewidth]{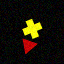}\par
\scriptsize{\example{A red triangle is below a cross.}}\par\ \par
\textcolor{green}{True}
\end{minipage}\,\begin{minipage}{0.11\linewidth}
\centering
\phantom{training}\par
\includegraphics[width=\linewidth]{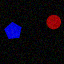}\par
\scriptsize{\example{A blue shape is to the left of a circle.}}\par
\textcolor{green}{True}
\end{minipage}$\,\Rightarrow\,$\begin{minipage}{0.11\linewidth}
\centering
evaluation\phantom{g}\par
\includegraphics[width=\linewidth]{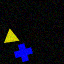}\par
\scriptsize{\example{A triangle is below a blue cross.}}\par
\textcolor{orange}{???}
\end{minipage}
\end{center}
\vspace{-0.2cm}
\caption{\label{figure:task}
A few example instances for the ICA task, which require the system to judge whether a caption is consistent with an image. Moreover, the two sequences outline the kind of generalization capability we are evaluating for, where previously seen concepts have to be recombined in novel ways.}
\end{figure*}

Deep learning methods have had a major impact on research in natural language processing and raised performance substantially in many of the standard evaluations. Moreover, multimodal tasks like image captioning \cite{Karpathy2015} or visual question answering (VQA) \cite{Antol2015} can now be tackled with great success. Such systems seem to solve the problems entirely on a sub-symbolic level, based only on raw image (and text) input, whereas previous approaches required a hand-crafted combination of various higher-level components.

There is, however, concern about how deep neural networks learn to solve such tasks. Investigations for image recognition \cite{Szegedy2014,Nguyen2015,Zhang2017} have shown surprising behavior very different from what would be expected of their \emph{``surpassing human-level performance''} \cite{He2015}. Deep networks for language tasks may exhibit similarly odd behavior \cite{Sproat2016,Arthur2016}. Moreover, it was recently found that datasets -- for instance in VQA -- contain various unexpected biases and peculiarities, which systems can exploit to answer complex questions, sometimes even without looking at the image at all \cite{Goyal2016,Agrawal2016,Zhang2016}. Such results cast doubt on whether deep learning systems actually acquire appropriate generalizations. However, given the recursive nature of language and the potentially enormous problem space of VQA and similar tasks, acquiring the ability for reliable generalization will eventually be essential.

A more theoretical issue is the ability of network architectures, in principle, to learn certain classes of structure. For instance, it has been shown that LSTMs possess the ability to handle long-range dependencies \cite{Hochreiter1997,Gers2001}. However, the formal experiments that have been done along such lines are limited, particularly in the multimodal domain of vision and language. While recent work indicates that the information encoded in image embeddings might be rich enough for good captioning results, it is an open question whether current architectures are able, \emph{in principle}, to combine visual information effectively to handle the full range of linguistic constructions.

This paper introduces a new test methodology for multimodal deep learning models. \shapeworld\ is a framework for specifying VQA-style datasets. An instance here consists of an image and a caption, and the evaluated model has to decide about their agreement, hence a form of yes/no question answering which we call \emph{image caption agreement} (ICA). Figure \ref{figure:task} illustrates the task and nature of the data with some example instances.

A \shapeworld\ dataset differs from standard VQA evaluation datasets in three main ways. Firstly, a \shapeworld\ dataset defines a \emph{process for generating artificial data} consisting of abstract colored shapes, which is randomly sampled during training/testing according to constraints specified by the experimenter. Secondly, the evaluation focus is on linguistic understanding capabilities of the type investigated by \emph{formal semantics}. The visual complexity and open-class vocabulary size is reduced to a minimum, while potentially allowing indefinitely complex syntactic constructions. Finally, the distribution of the evaluation data is \emph{deliberately kept different} from the training distribution. Controlled data generation enables us to introduce previously unseen instance configurations during evaluation, which require the system to recombine learned concepts to be able to understand these novel instances (see figure \ref{figure:task}) -- hence a form of \emph{zero-shot learning}. We think of the \shapeworld\ tasks as unit-testing multimodal systems for specific linguistic generalization capabilities, in a similar way to the bAbI tasks \cite{Weston2015} for text-only understanding.

We also present results for various VQA models on four \shapeworld\ datasets targeting different multimodal language understanding abilities. The artificial data allows for a detailed analysis of the models' strengths and weaknesses, and reveals unexpected shortcomings. As such, it offers a significantly different and interesting resource to complement standard evaluation practice. By exposing problematic instance patterns where these systems fail (e.g.\ spatial relations), and by providing \emph{a configurable, extensible testbed for systematic, detailed and comparable evaluation}, we hope to stimulate progress in the field.

\section{Related work}

With the increasing popularity of deep learning approaches, artificial data of various kinds is again seen as a valuable tool in experimentation. Recently, the simulation paradigm has been argued to be a promising driver for artificial intelligence research \cite{Kiela2016}. Various platforms following this paradigm have been released, mostly aimed at reinforcement learning: the Arcade Learning Environment / Atari 2600 games \cite{Bellemare2013}, OpenAI Gym \cite{OpenAI2016}, DeepMind Lab \cite{DeepMindLab2016}, Project Malmo \cite{Malmo2016}, to name a few of the most popular. An important advantage of simulated data is its infinite availability, particularly in light of the need of many deep learning models for huge amounts of data. Automatically generating data greatly reduces the cost, time and human effort. Moreover, it allows researchers to focus on specific problem situations, isolated from a noisy and complex real-world environment.

When focusing on language tasks, the simulation paradigm faces the problem that interesting language generation is a difficult task in its own right, and that the difficulty increases with the complexity of the underlying world. The bAbI tasks \cite{Weston2015} are generated by internally simulating a short scene and extracting a few simple sentences from it. A similar approach is taken by \newcite{Narasimhan2015}, but here the simulation is more complex, comprising a text-based role-playing game. The MazeBase game environment \cite{Sukhbaatar2015} uses language as a mean to represent the game world. However, the descriptions are in an abstract, formulaic format, and the focus of the simulation is much more on the planning than the language component. The long-term research proposal of \newcite{Mikolov2015} also simulates a world where an agent learns to solve tasks by communication with a teacher module. At least for a start, this module is supposed to be scripted to automatically generate appropriate responses, given its internal knowledge of the world state.

Automatically generated data is common for tasks specifically focusing on the ability to efficiently process data of a certain formal structure. Here, data is deliberately stripped of any real-world connection to create an abstract capability check. Recent work in the context of deep learning has investigated sequence patterns \cite{Joulin2015}, combinatorial problems \cite{Vinyals2015}, or executing programming language code \cite{Zaremba2014}, amongst others. This kind of task is particularly common for neural network models (see, for instance, \newcite{Bengio1994} more than twenty years ago). The reason for interest in abstract capability checks is that the learning process and decisions of deep networks are more difficult to interpret than shallower machine learning methods. \newcite{Bowman2015} and \newcite{Sorodoc2016} are more similar to our work in focusing on specific linguistic aspects. Both generate artificial data automatically based on abstract models for tasks targeting logical semantics and quantifiers, respectively.

The multimodal tasks of image captioning and VQA are closely related to our evaluation goal, but usually consist of \emph{``repurposed''} real-world photos and human-written descriptions.\footnote{Although we contrast such ``real-world'' data with artificial simulations, it should be clear that this is very unlike the visual experience of an entity situated in the real world.} However, there have been experiments in which parts of the data are artificial and/or generated automatically, for instance, automatic question generation from annotation \cite{Ren2015} or systematic modification of captions \cite{Hodosh2016}. Abstract Clipart scenes have been used for image captioning \cite{Zitnick2016,Zitnick2013b} and to balance existing VQA datasets \cite{Zhang2016}. Most similar to the \shapeworld\ framework is the CLEVR dataset \cite{Johnson2017}. It contains images of rendered abstract 3-dimensional scenes and complex questions generated from a variety of templates. As with our work, they propose that their artificial dataset \emph{complements} evaluation on real-world VQA datasets.

Our own work is based on automatically generated, fully artificial data. This data is not specifically designed to address only a single structural problem, but is a testbed able to cover a whole range of linguistic phenomena. In fact, our generation system closely resembles classical work in formal semantics, where a statement corresponds to a logical expression which can be evaluated against an abstract world model \cite{Montague1970}. We utilize semantic representations based on \textit{Minimal Recursion Semantics} \cite{Copestake2005} and broad-coverage, grammar-based realization driven by the \textit{English Resource Grammar} \cite{Flickinger2000} to make the internal world model compatible with language. However, while \shapeworld\ uses these abstract representations internally, the \emph{external} representation presented to the system under evaluation does not involve any abstract formalization of visual and textual input. It nevertheless presents the intended problems clearly, without any uncontrolled noise, biases or hidden correlations, which can obfuscate results when using real-world images and text \cite{Goyal2016,Agrawal2016}.

\section{The \shapeworld\ framework}

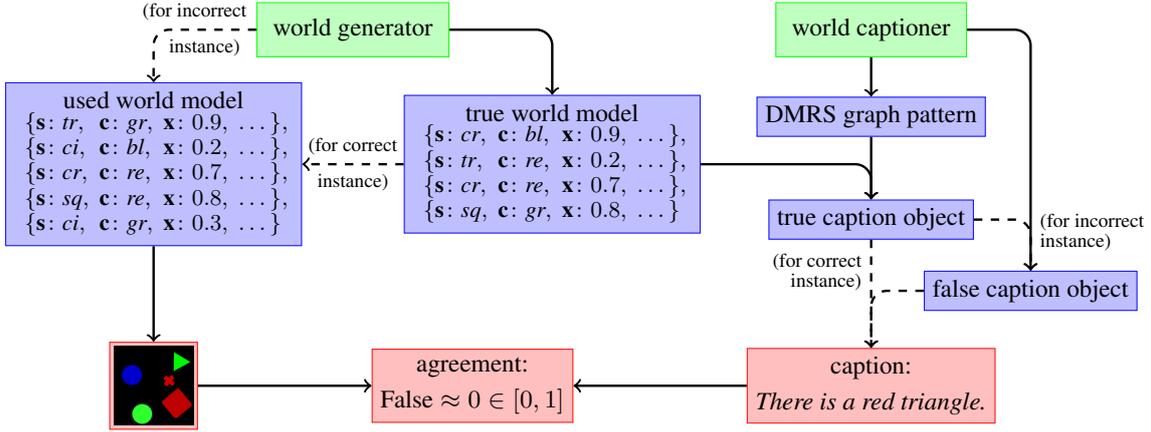
\begin{figure*}
\centering\resizebox{0.95\linewidth}{!}{
\small
\begin{tikzpicture}
\node[rectangle,fill=green!25,draw=green!90,inner sep=0.2cm] (generator) at (2.5, 4.5) {world generator};
\node[rectangle,fill=green!25,draw=green!90,inner sep=0.2cm] (captioner) at (9, 4.5) {world captioner};
\node[rectangle,fill=blue!25,draw=blue!90] (model2) at (0, 2.8) [align=center] {used world model\\
\resizebox{3.5cm}{!}{
\begin{math}
\footnotesize
\setlength{\arraycolsep}{0.07cm}
\begin{array}{llll}
\{\textbf{s}\colon \textit{tr}, & \textbf{c}\colon \textit{gr}, & \textbf{x}\colon 0.9, & \dots\},\\
\{\textbf{s}\colon \textit{ci}, & \textbf{c}\colon \textit{bl}, & \textbf{x}\colon 0.2, & \dots\},\\
\{\textbf{s}\colon \textit{cr}, & \textbf{c}\colon \textit{re}, & \textbf{x}\colon 0.7, & \dots\},\\
\{\textbf{s}\colon \textit{sq}, & \textbf{c}\colon \textit{re}, & \textbf{x}\colon 0.8, & \dots\},\\
\{\textbf{s}\colon \textit{ci}, & \textbf{c}\colon \textit{gr}, & \textbf{x}\colon 0.3, & \dots\}
\end{array}
\end{math}}};
\node[rectangle,fill=blue!25,draw=blue!90] (model1) at (5, 2.8) [align=center] {true world model\\
\resizebox{3.5cm}{!}{
\begin{math}
\footnotesize
\setlength{\arraycolsep}{0.07cm}
\begin{array}{lllll}
\{\textbf{s}\colon \textit{cr}, & \textbf{c}\colon \textit{bl}, & \textbf{x}\colon 0.9, & \dots\},\\
\{\textbf{s}\colon \textit{tr}, & \textbf{c}\colon \textit{re}, & \textbf{x}\colon 0.2, & \dots\},\\
\{\textbf{s}\colon \textit{cr}, & \textbf{c}\colon \textit{re}, & \textbf{x}\colon 0.7, & \dots\},\\
\{\textbf{s}\colon \textit{sq}, & \textbf{c}\colon \textit{gr}, & \textbf{x}\colon 0.8, & \dots\}
\end{array}
\end{math}}};
\node[rectangle,fill=blue!25,draw=blue!90] (pattern) at (9, 3.4) [align=center] {DMRS graph pattern};
\node[rectangle,fill=blue!25,draw=blue!90] (caption1) at (9, 2.1) [align=center] {true caption object};
\node[rectangle,fill=blue!25,draw=blue!90] (caption2) at (11, 1.2) [align=center] {false caption object};
\node[rectangle,fill=red!25,draw=red!90,inner sep=0.05cm] (image) at (0, 0) [align=center] {\includegraphics[width=1cm]{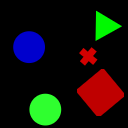}};
\node[rectangle,fill=red!25,draw=red!90] (agreement) at (4, 0) [align=center] {agreement:\\[0.1cm]$\text{False} \approx 0 \in [0, 1]$};
\node[rectangle,fill=red!25,draw=red!90] (text) at (9, 0) [align=center] {caption:\\[0.1cm]\example{There is a red triangle.}};
\draw[->,line width=0.03cm,rounded corners=0.2cm] (generator) -| (model1);
\draw[->,line width=0.03cm,rounded corners=0.2cm,dashed] (generator) -| node[above,pos=0.3] {\scriptsize{(for incorrect}} node[below,pos=0.3] {\scriptsize{\quad instance)}} (model2);
\draw[->,line width=0.03cm,dashed] (model1) -- node[above] {\scriptsize{(for correct}} node[below] {\scriptsize{instance)}} (model2);
\draw[->,line width=0.03cm] (captioner) -- (pattern);
\draw[->,line width=0.03cm,rounded corners=0.2cm] (pattern) -- (caption1);
\draw[->,line width=0.03cm] (model2) -- (image);
\draw[->,line width=0.03cm,rounded corners=0.2cm] (model1) -| (caption1);
\draw[->,line width=0.03cm,rounded corners=0.2cm] (captioner) -| node[right,pos=0.92,align=left] {\scriptsize{(for incorrect}\\[-0.1cm]\scriptsize{instance)}} (caption2);
\draw[->,line width=0.03cm,rounded corners=0.2cm,dashed] (caption1) -| (caption2);
\draw[->,line width=0.03cm,dashed] (caption1) -- node[left,pos=0.3,align=right] {\scriptsize{(for correct}\\[-0.1cm]\scriptsize{instance)}} (text);
\draw[->,line width=0.03cm,rounded corners=0.2cm,dashed] (caption2) -| (text);
\draw[->,line width=0.03cm] (image) -- (agreement);
\draw[->,line width=0.03cm] (text) -- (agreement);
\end{tikzpicture}}
\caption{\label{figure:generation}
The generation process for ICA data, showing the alternative pathways depending on whether a correct instance (i.e., a true statement about the world) or an incorrect instance is to be generated.}
\end{figure*}

The \shapeworld\ framework\footnote{
The \shapeworld\ code is written in Python 3 and is available on GitHub (\githuburl). The generated data is returned as NumPy arrays, so that it is possible to integrate it into Python-based deep learning projects using common frameworks like TensorFlow, Theano, etc. In our experiments, we use TensorFlow and we provide the models in this paper as part of the package. For the internal DMRS-based caption generation, the Python package \textit{pydmrs} \cite{Copestake2016}, as well as a reduced version of the \textit{English Resource Grammar} \cite{Flickinger2000} and of Packard's \textit{Answer Constraint Engine} (\url{http://sweaglesw.org/linguistics/ace/}) is included.} is based on \newterm{microworlds} -- small and self-contained artificial scenarios -- which guide the data creation process. The \shapeworld\ microworlds simply consist of colored shapes. This closed-world domain allows for exhaustive coverage of the space of possible microworlds and associated captions. The vocabulary used has an emphasis on closed-class words -- the open-class vocabulary is currently far less than 100 words. In the following we explain the details of the data generation process inside the \shapeworld\ framework. A schematic illustration of the process is shown in figure \ref{figure:generation}.

\subsection{Image caption agreement task}
In this paper we focus on the task of \newterm{image caption agreement (ICA)}. The system to be evaluated is presented with an image and a natural language caption and has to decide whether they are consistent with each other.

Compared to the classic image captioning task, ICA emphasizes the understanding rather than the synthesis part of language use. We therefore avoid the problem of evaluating the \emph{appropriateness} of a caption. The setup allows us to control the content of both modalities and consequently force a system to cope with difficult types of captions while obtaining a clear indicator of successful understanding. Although very similar to the VQA setup (i.e., yes/no questions), it neither requires the evaluated model to generate answers nor to rephrase the problem to fit it into a classification task of some sort -- for instance, over the 1000/3000 most common answers, as is common practice recently \cite{Lu2016,Fukui2016}. ICA most closely corresponds to the work of \newcite{Jabri2016}, who present VQA as a binary classification of image-question-answer triples.

One further motivation for the task is that human performance could be measured using the same setup.  We would expect close-to-perfect human performance on the tasks described here, assuming time is not tightly constrained. Interesting comparisons are potentially possible where human performance depends on presentation: e.g., quantifiers such as \example{most} \cite{Pietroski2009}. However, we will not discuss this further in the current paper.

\begin{figure*}[!htb]
\centering
\includegraphics[width=0.9\linewidth]{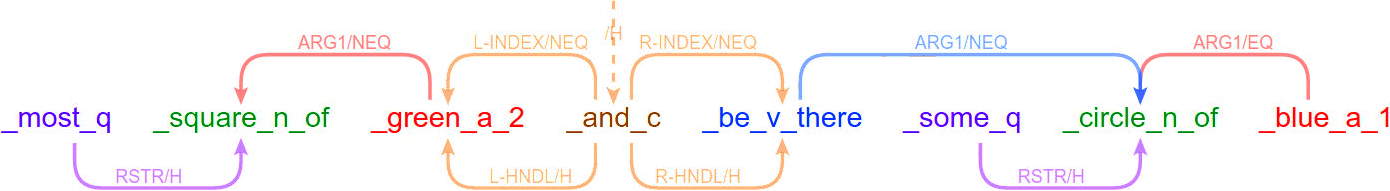}\par
\footnotesize{\example{Most squares are green and there are some circles which are blue.}}\par
\vspace{0.4cm}
\resizebox{0.9\linewidth}{!}{
\begin{math}
\approx
\left[
\frac
{\#\{ s_1 \in \text{World} \colon \text{square}(s_1.\text{shape}) \,\wedge\, \text{green}(s_1.\text{color}) \}}
{\#\{ s_2 \in \text{World} \colon \text{square}(s_2.\text{shape}) \}}
> \frac{1}{2}
\right]
\;\wedge\;
\Big[
\exists s_3 \in \text{World}  \colon \text{circle}(s_3.\text{shape}) \wedge \text{blue}(s_3.\text{color})
\Big]
\end{math}
}
\caption{\label{figure:dmrs}
An example of a DMRS graph corresponding to a more complex caption, with compositional components colored. The logical formula gives the formal semantic interpretation over a world model.}
\end{figure*}

\subsection{World and image generation}
At the core of each microworld instance lies an abstract \newterm{world model}. The internal representation of a microworld is simply a list of \newterm{entities}, given as records containing their \newterm{primary attributes}, such as \textit{position}, \textit{shape}, \textit{color}, which are considered to be high-level semantic aspects reflected in captions. In addition, an entity has \newterm{secondary attributes} and methods which control, for instance, details of visual appearance, visual noise infusion, or the collision-free placement of entities. Importantly, all these ways of infusing noise can be controlled as well, which is useful particularly since noise is often seen as important for successful training of deep models.

The \newterm{generator module} automatically generates a world model by randomly sampling all these attributes from a set of available values. Both these values and other aspects of the generation process can be specified and adjusted appropriately for each dataset. The internal abstract representation is then used as a basis to extract a concrete microworld instance consisting of image and caption. The image (of size 64$\times$64 in this work) is just a straightforward visualization of the world model. The table below gives an overview of the primary and secondary attributes, together with the value ranges and sampling details used for experiments in this paper (``distortion'' here means width divided by height for rectangles and ellipses).

\begin{center}
\small
\begin{tabular}{|p{1.5cm}|p{4.5cm}|}
\hline
\ \newline shape & $\text{choose}(\example{square}, \example{rectangle}, \example{triangle},\newline\hspace*{0.1cm} \example{pentagon}, \example{cross}, \example{circle}, \example{semicircle},\newline\hspace*{0.1cm} \example{ellipse})$\\
\vspace{-0.25em} color & $\text{choose}(\example{red}, \example{green}, \example{blue}, \example{yellow},\newline\hspace*{0.1cm} \example{magenta}, \example{cyan}, \example{white})$\\
location & $\text{uniform}(a=(0,0),\: b=(64,64))$\\
\hline
object size & $\text{uniform}(a=0.15,\: b=0.3)$\\
distortion & $\text{uniform}(a=2.0,\: b=3.0)$\\
rotation & $\text{uniform}(a=0.0,\: b=1.0)$\\
shade & $\text{trunc\_normal}(\mu=0.0,\: \sigma=0.5)$\\
\hline
pixel noise & $\text{trunc\_normal}(\mu=0.0,\: \sigma=0.1)$\\
\hline
\end{tabular}
\end{center}

\subsection{Caption generation}

\begin{figure*}[!htb]
\small
\begin{minipage}[t]{0.24\linewidth}
\centering
\textsc{\normalsize\phantom{Q}OneShape\phantom{Q}}\par
\includegraphics[width=0.48\linewidth]{oneshape1.png}\hspace{0.05cm}\includegraphics[width=0.48\linewidth]{oneshape2.png}
\begin{flushleft}
\enskip\textbullet\enskip \example{There is a blue rectangle.}\par
\enskip\textbullet\enskip \example{There is a triangle.}\par
\enskip\textbullet\enskip \example{There is a yellow shape.}\par
\vspace{0.2cm}
\textbf{Training combinations:}\par
\enskip\textbullet\enskip 50 combinations\par
\vspace{0.1cm}
\textbf{Validation combinations:}\par
\enskip\textbullet\enskip \example{red square}, \example{green\\\qquad triangle}, \example{blue circle}\par
\vspace{0.1cm}
\textbf{Test combinations:}\par
\enskip\textbullet\enskip \example{yellow rectangle}, \example{cyan\\\qquad ellipse}, \example{magenta cross}
\end{flushleft}
\end{minipage}
\begin{minipage}[t]{0.24\linewidth}
\centering
\textsc{\normalsize\phantom{Q}MultiShape\phantom{Q}}\par
\includegraphics[width=0.48\linewidth]{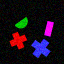}\hspace{0.05cm}\includegraphics[width=0.48\linewidth]{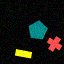}
\begin{flushleft}
\enskip\textbullet\enskip \example{There is a magenta\\\qquad semicircle.}\par
\enskip\textbullet\enskip \example{There is a pentagon.}\par
\enskip\textbullet\enskip \example{There is a cyan shape.}\par
\vspace{0.1cm}
\enskip(Same as for \textsc{OneShape})\par
\vspace{0.2cm}
\textbf{Number of objects}\par
\enskip\textbullet\enskip Training: 1, 2, 3, 5\par
\enskip\textbullet\enskip Validation: 4\par
\enskip\textbullet\enskip Testing: 6
\end{flushleft}
\end{minipage}
\begin{minipage}[t]{0.24\linewidth}
\centering
\textsc{\normalsize\phantom{Q}Spatial\phantom{Q}}\par
\includegraphics[width=0.48\linewidth]{spatial1.png}\hspace{0.05cm}\includegraphics[width=0.48\linewidth]{spatial2.png}
\begin{flushleft}
\enskip\textbullet\enskip \example{A red circle is to the left\\\qquad of a cyan semicircle.}\par
\enskip\textbullet\enskip \example{A white circle is above a\\\qquad pentagon.}\par
\enskip\textbullet\enskip \example{A red shape is to the right\\\qquad of a green triangle.}\par
\enskip\textbullet\enskip \example{A shape is below a cross.}\par
\vspace{0.2cm}
\textbf{Training / validation / test combinations:}\par
\enskip\textbullet\enskip Same as for \textsc{OneShape}
\end{flushleft}
\end{minipage}
\begin{minipage}[t]{0.24\linewidth}
\centering
\textsc{\normalsize Quantification}\par
\includegraphics[width=0.48\linewidth]{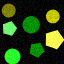}\hspace{0.05cm}\includegraphics[width=0.48\linewidth]{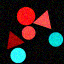}
\begin{flushleft}
\enskip\textbullet\enskip \example{The shape is green.}\par
\enskip\textbullet\enskip \example{Most shapes are\\\qquad rectangles.}\par
\enskip\textbullet\enskip \example{No shape is a red triangle.}\par
\enskip\textbullet\enskip \example{All triangles are green.}\par
\enskip\textbullet\enskip \example{Two blue shapes are\\\qquad pentagons.}\par
\vspace{0.2cm}
\textbf{Number of objects}\par
\enskip\textbullet\enskip Training: 3, 4, 5, 7\par
\enskip\textbullet\enskip Validation: 6\par
\enskip\textbullet\enskip Testing: 8
\end{flushleft}
\end{minipage}
\caption{\label{figure:datasets}
The four datasets used in this paper. For each of them, two example microworlds and a few captions are shown, as well as details about the structure of training, validation and test instances.}
\end{figure*}

We currently provide an implementation of the \shapeworld\   captioner interface using a grammar-based approach. More specifically, Dependency Minimal Recursion Semantics (DMRS) \cite{Copestake2016} is an abstract semantic graph representation designed for use with high-precision grammars, such as those distributed by the DELPH-IN consortium.\footnote{Although we currently use the English Resource Grammar \cite{Flickinger2000}, other DELPH-IN grammars use a compatible approach, so \shapeworld\ can easily be ported to other languages.}

A semantic representation like DMRS is particularly suited for the \shapeworld\ framework, since it essentially mirrors the internal world model and hence acts like a (partial) language-specific annotation. Here, noun nodes correspond to entities, adjective nodes add attributes, and verb phrase nodes/sub-graphs specify relations between entities. The semantics of words like \example{``square''} or \example{``red''} is interpreted as iteratively filtering a subset of agreeing entities, while transitive relations like \example{``to the left of''} act similarly on pairs of entity sets, and quantifiers compare the cardinality of two entity sets. Below an example of a DMRS semantic graph with its compositional components colored:
\begin{center}
\begin{minipage}{0.9\linewidth}
\centering
\includegraphics[width=\linewidth]{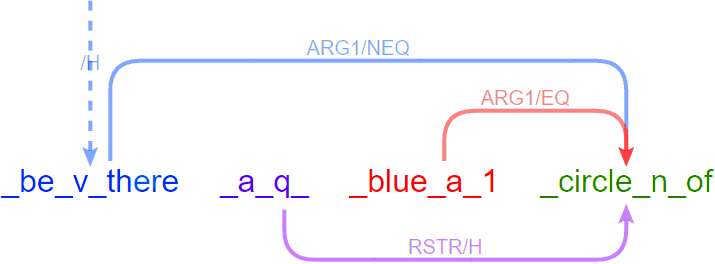}\par
\footnotesize{\example{There is a blue circle.}}
\end{minipage}
\end{center}

\emph{Compositionality} of the semantic representation is a useful property and an important reason for our use of DMRS. Given compositionality, it is enough to specify the semantics of words -- or, more precisely, of the linguistic atoms in the \shapeworld\ context, which potentially are sub-graphs with multiple nodes and inner link structure -- to be able to obtain the corresponding semantics of composed sub-graphs, and so generate a wide range of different captions.

Figure \ref{figure:dmrs} shows an example of a more complex compositional caption, which contains the DMRS graph above as sub-graph. It also illustrates how various details are automatically inferred by the English Resource Grammar, including number-agreement between subject and verb, and between quantifier and noun, and realization of an adjective as relative clause. This greatly facilitates the generation of a combinatorially large amount of captions and makes the DMRS graph patterns reusable. Finally, figure \ref{figure:dmrs} gives a formal semantic interpretation of the caption meaning as logical formula over a world model. This indicates how the agreement of a caption with a microworld is computed in the \shapeworld\ framework.

Similar to the generator module, the \newterm{captioner module} randomly samples from a set of dataset-specific \newterm{DMRS graph patterns}, which are then applied to a world model to construct an \emph{agreeing} \newterm{caption object} (see figure \ref{figure:generation}). The DMRS graph can be turned into an MRS representation, from which a corresponding English sentence can be generated with a bi-directional grammar like the \textit{English Resource Grammar} and a parser-generator like Packard's \textit{Answer Constraint Engine}.

The captioner module's ability to check whether another world model would agree with the semantics of this caption is important for the generation of negative instances, i.e., caption/microworld pairs that do not agree. These instances are obtained by either sampling a second, \emph{false} world model, or by producing a \emph{false} caption object via modification of the agreeing caption. In either case, the system ensures that \emph{false} microworld and caption object do not accidentally agree.

\subsection{Training and testing on \shapeworld\ datasets}
Since \shapeworld\ datasets are actually data generation processes, training and evaluation work differently from classic datasets. Where usually one has a fixed set of instances, here models are trained and tested on a fixed set of higher-level \newterm{generator configuration constraints}. In particular, the constraints for evaluation differ from the training constraints, hence requiring true generalization abilities. For instance, a certain shape-color-combination, a specific number of objects, a spatial location or a caption type can be held-out and never generated during training, such that concepts need to be recombined at test time. It is thus possible for a system to achieve optimal performance during training, but completely fail the evaluation.

Another important property of the \shapeworld\ datasets, particularly for future extensions, is their \emph{compositionality}. Instead of having to define a dataset from scratch every time, we can specify \newterm{atomic datasets} and then combine them in a \newterm{mixer dataset}, which tests for various different aspects of multimodal language understanding simultaneously. Reusability in fact applies even further down in the component hierarchy. For instance, we use the same generic world generator module for all four datasets. This is also useful for caption generation where, for instance, a logical combinator dataset can reuse different world captioner modules to generate simple statements which then are merged by logical connectives.

\section{Experiments}

\begin{figure*}[!htb]
\centering\resizebox{0.95\linewidth}{!}{
\small
\begin{tikzpicture}[node distance=0.5cm]
\node[rectangle,fill=red!25,draw=red!90,inner sep=0.05cm] (image) [align=center] {\includegraphics[width=1cm]{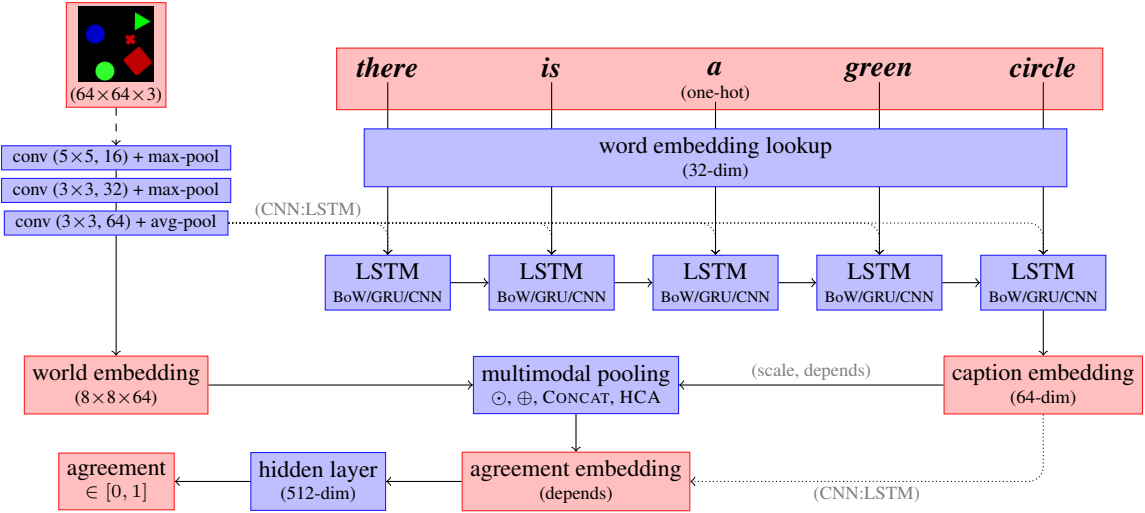}\\[-0.1cm]\scriptsize{(64$\times$64$\times$3)}};
\node[rectangle,fill=blue!25,draw=blue!90,inner sep=0.05cm] (conv1) [below=of image] {\scriptsize{\hspace*{0.1cm}conv (5$\times$5, 16) + max-pool\hspace*{0.1cm}}};
\node[rectangle,fill=blue!25,draw=blue!90,inner sep=0.05cm] (conv2) [below=0.1cm of conv1] {\scriptsize{\hspace*{0.1cm}conv (3$\times$3, 32) + max-pool\hspace*{0.1cm}}};
\node[rectangle,fill=blue!25,draw=blue!90,inner sep=0.05cm] (conv3) [below=0.1cm of conv2] {\scriptsize{\hspace*{0.1cm}conv (3$\times$3, 64) + avg-pool\hspace*{0.1cm}}};
\node[rectangle,fill=red!25,draw=red!90] (world) [below=1.6cm of conv3,align=center] {world embedding\\[-0.05cm]\scriptsize{(8$\times$8$\times$64)}};
\draw[->,dashed] (image) -- (conv1);
\draw[-] (conv1) -- (conv2);
\draw[-] (conv2) -- (conv3);
\draw[->] (conv3) edge (world);

\node[rectangle,fill=blue!25,draw=blue!90] (pooling) [right=3.5cm of world,align=center] {multimodal pooling\\[-0.05cm]\scriptsize{$\odot$, $\oplus$, \textsc{Concat}, \textsc{HCA}}};
\node[rectangle,fill=red!25,draw=red!90] (caption) [right=3.5cm of pooling,align=center] {caption embedding\\[-0.05cm]\scriptsize{(64-dim)}};
\draw[->] (world) -- (pooling);
\draw[->] (caption) -- node[above=-0.05cm] {\textcolor{gray}{\scriptsize{(scale, depends)}}} (pooling);

\node[rectangle,fill=blue!25,draw=blue!90] (lstm5) [above=0.6cm of caption,align=center] {LSTM\\[-0.05cm]\tiny{BoW/GRU/CNN}};
\node[rectangle,fill=blue!25,draw=blue!90] (lstm4) [left=of lstm5,align=center] {LSTM\\[-0.05cm]\tiny{BoW/GRU/CNN}};
\node[rectangle,fill=blue!25,draw=blue!90] (lstm3) [left=of lstm4,align=center] {LSTM\\[-0.05cm]\tiny{BoW/GRU/CNN}};
\node[rectangle,fill=blue!25,draw=blue!90] (lstm2) [left=of lstm3,align=center] {LSTM\\[-0.05cm]\tiny{BoW/GRU/CNN}};
\node[rectangle,fill=blue!25,draw=blue!90] (lstm1) [left=of lstm2,align=center] {LSTM\\[-0.05cm]\tiny{BoW/GRU/CNN}};

\node[inner sep=0cm] (t1) [above=2.3cm of lstm1] {\phantom{g\textit{\textbf{\normalsize there}}g}};
\node[inner sep=0cm] (t3) [above=2.04cm of lstm3,align=center] {\phantom{\textit{\textbf{\normalsize a}}}\\[-0.05cm] \phantom{\scriptsize{(one-hot)}}};
\node[inner sep=0cm] (t5) [above=2.3cm of lstm5] {\phantom{gh\textit{\textbf{\normalsize circle}}g}};
\node[rectangle,fill=red!25,draw=red!90,fit={(t1) (t3) (t5)}] {};

\node[inner sep=0cm] (w1) [above=2.3cm of lstm1] {\phantom{g}\textit{\textbf{\normalsize there}}\phantom{g}};
\node[inner sep=0cm] (w2) [above=2.3cm of lstm2] {\phantom{gh}\textit{\textbf{\normalsize is}}\phantom{gh}};
\node[inner sep=0cm] (w3) [above=2.04cm of lstm3,align=center] {\phantom{gh}\textit{\textbf{\normalsize a}}\phantom{gh}\\[-0.05cm]\scriptsize{(one-hot)}};
\node[inner sep=0cm] (w4) [above=2.3cm of lstm4] {\phantom{h}\textit{\textbf{\normalsize green}}\phantom{h}};
\node[inner sep=0cm] (w5) [above=2.3cm of lstm5] {\phantom{g}\textit{\textbf{\normalsize circle}}\phantom{g}};

\draw[->] (w1) -- (lstm1);
\draw[->] (w2) -- (lstm2);
\draw[->] (w3) -- (lstm3);
\draw[->] (w4) -- (lstm4);
\draw[->] (w5) -- (lstm5);

\node[rectangle,fill=blue!25,draw=blue!90] (lookup) [above=0.9cm of lstm3,align=center] {\hspace*{3cm}word embedding lookup\hspace*{3cm}\\[-0.05cm]\scriptsize{(32-dim)}};

\draw[->] (lstm1) -- (lstm2);
\draw[->] (lstm2) -- (lstm3);
\draw[->] (lstm3) -- (lstm4);
\draw[->] (lstm4) -- (lstm5);
\draw[->] (lstm5) -- (caption); 
\draw[->,densely dotted,rounded corners=0.2cm] (conv3) -| node[above=-0.05cm,pos=0.25] {\textcolor{gray}{\scriptsize{(CNN:LSTM)}}} (lstm1);
\draw[->,densely dotted,rounded corners=0.2cm] (conv3) -| (lstm2);
\draw[->,densely dotted,rounded corners=0.2cm] (conv3) -| (lstm3);
\draw[->,densely dotted,rounded corners=0.2cm] (conv3) -| (lstm4);
\draw[->,densely dotted,rounded corners=0.2cm] (conv3) -| (lstm5);

\node[rectangle,fill=red!25,draw=red!90] (embedding) [below=of pooling,align=center] {agreement embedding\\[-0.05cm]\scriptsize{(depends)}};
\node[rectangle,fill=blue!25,draw=blue!90] (hidden) [left=1cm of embedding,align=center] {hidden layer\\[-0.05cm]\scriptsize{(512-dim)}};
\node[rectangle,fill=red!25,draw=red!90] (agreement) [left=1cm of hidden,align=center] {agreement\\[-0.05cm]\scriptsize{$\in [0,1]$}};
\draw[->] (pooling) -- (embedding);
\draw[->] (embedding) -- (hidden);
\draw[->] (hidden) -- (agreement);
\draw[->,densely dotted,rounded corners=0.2cm] (caption) |-  node[below=-0.05cm,pos=0.75] {\textcolor{gray}{\scriptsize{(CNN:LSTM)}}} (embedding);
\end{tikzpicture}
}
\vspace{-0.2cm}
\caption{\label{figure:architecture}
Overview of the VQA architectures we evaluate, with details on model parameters.}
\end{figure*}

\subsection{Datasets}
In this paper we look at four datasets, each designed to investigate an aspect of the capability to understand language in a multimodal setup. Figure \ref{figure:datasets} gives further information about these datasets. Note that since we first sample a microworld model and subsequently a caption, we cannot always easily control the generation process to sample each possible caption \emph{perfectly} uniformly. This is in particular the case when focusing on more specific captions which might not apply to a microworld and hence require resampling.

\subsection{Network architecture}
We evaluate several multimodal deep neural network architectures that were recently proposed for VQA \cite{Goyal2016,Agrawal2016,Jabri2016,Antol2015,Ren2015}. Figure \ref{figure:architecture} shows the general architecture underlying all of these models. Implementations in TensorFlow, adapted for the ICA task, are included as part of the GitHub repository\footnote{\githuburl}. Each model is trained end-to-end on the task, including the CNN module and the word embeddings, as opposed to using pre-trained, general-purpose versions. We train for 5000 iterations\footnote{We tracked the validation performance and found that learning essentially plateaus after at most half the iterations.} with a batch size of 128, using Adam optimization \cite{Kingma2014} with learning rate 0.001.

\textit{LSTM-only} and \textit{CNN-only} are simple unimodal baselines. \textit{CNN+\{BoW,LSTM,GRU\}:Mult} obtain the caption embedding via BoW, LSTM or GRU, respectively, then fuse visual and textual information via pointwise multiplication. \textit{CNN+LSTM:Add} and \textit{CNN+LSTM:Concat}, i.e., pointwise addition and concatenation, are alternative basic ways of combining image and caption embeddings. Instead of concatenating the image embedding with the output of the LSTM, in \textit{CNN:LSTM} it is concatenated with each word embedding before being processed by the LSTM. Finally, hierarchical co-attention \cite{Lu2016} combines visual information on word-, phrase- and sentence-level with the language input, which is processed by a CNN. \textit{CNN+CNN:HCA-\{par,alt\}} implements this approach with the two proposed co-attention mechanisms, \textit{parallel} and \textit{alternating}.

In the near future, we plan to also adapt the technique of \textit{multimodal compact bilinear pooling} \cite{Fukui2016}, \textit{neural module networks} \cite{Andreas2016b,Andreas2016a} and potentially also \textit{relation networks} \cite{Raposo2017} to the ICA task, and upload implementations to the GitHub repository.

\subsection{Results}

\begin{figure*}
\small
\begin{center}
\begin{tabular}{|l|c|c|c|c|c|c|}
\hline
Dataset configuration & LSTM-only & CNN+LSTM:Mult & CNN+CNN:HCA-par & CNN+CNN:HCA-alt \\
\hline

\textsc{OneShape} &
51 / 46 / 50 &
81 / 70 / 66 &
90 / 77 / 78 &
\textbf{92 / 81 / 77} \\
\hline
\enskip C: no hypernyms &
90 / 70 / 100 \muchbetter &
95 / 64 / 57 \inconsistent &
98 / 71 / 73 \inconsistent &
97 / 68 / 66 \inconsistent \\
\enskip C: only hypernyms &
100 / 100 / 100 \muchbetter &
52 / 34 / 30 \muchworse &
96 / 78 / 82 \better &
95 / 75 / 73 \inconsistent \\
\enskip I: changed shape &
6 / 5 / 7 \muchworse &
70 / 81 / 82 \inconsistent &
60 / 63 / 58 \muchworse &
73 / 78 / 78 \worse \\
\enskip I: changed color &
8 / 15 / 0 \muchworse &
100 / 100 / 99 \muchbetter &
100 / 92 / 96 \muchbetter &
100 / 97 / 89 \muchbetter \\
\enskip I: changed both &
7 / 5 / 6 \muchworse &
96 / 97 / 98 \muchbetter &
87 / 85 / 84 \inconsistent &
93 / 92 / 89 \better \\
\hline

\textsc{MultiShape} &
62 / 67 / 67 &
\textbf{72 / 71 / 72} &
\textbf{72 / 71 / 69} &
71 / 68 / 68 \\
\hline
\enskip correct instances &
48 / 49 / 50 \muchworse &
76 / 64 / 54 \inconsistent &
81 / 68 / 65 \inconsistent &
71 / 59 / 53 \worse \\
\enskip I: random attr. &
58 / 63 / 68 \worse &
67 / 74 / 79 \inconsistent &
64 / 67 / 68 \worse &
70 / 73 / 78 \better \\
\enskip I: random existing attr. &
100 / 100 / 100 \muchbetter &
78 / 86 / 95 \muchbetter &
55 / 71 / 79 \inconsistent &
72 / 87 / 95 \muchbetter \\
\hline

\textsc{Spatial} &
52 / 51 / 50 &
57 / 52 / 54 &
\textbf{63 / 65 / 64} &
54 / 52 / 55 \\
\hline
\enskip C: no hypernyms &
85 / 85 / 69 \muchbetter &
45 / 44 / 41 \worse &
83 / 83 / 86 \muchbetter &
92 / 62 / 100 \muchbetter \\
\enskip C: only hypernyms &
95 / 95 / 97 \muchbetter &
4 / 6 / 4 \muchworse &
60 / 59 / 65 \worse &
49 / 40 / 52 \worse \\
\enskip I: swapped direction &
11 / 13 / 16 \muchworse &
98 / 97 / 98 \muchbetter &
36 / 39 / 30 \muchworse &
50 / 61 / 47 \inconsistent \\
\enskip I: object random attr. &
15 / 12 / 16 \muchworse &
88 / 88 / 91 \muchbetter &
69 / 68 / 68 \better &
63 / 66 / 60 \better \\
\enskip I: subject random attr. &
13 / 12 / 17 \muchworse &
87 / 88 / 89 \muchbetter &
69 / 71 / 70 \better &
61 / 64 / 56 \better \\
\hline

\textsc{Quantification} &
57 / 57 / 56 &
56 / 56 / 58 &
\textbf{76 / 77 / 78} &
\textbf{74 / 77 / 78} \\
\hline
\enskip correct instances &
23 / 22 / 18 \muchworse &
25 / 30 / 26 \muchworse &
74 / 71 / 72 \worse &
70 / 71 / 75 \worse \\
\enskip incorrect instances &
94 / 93 / 93 \muchbetter &
88 / 90 / 88 \muchbetter &
81 / 83 / 88 \better &
78 / 82 / 82 \better \\
\enskip instances with \example{no} &
52 / 51 / 48 \worse &
61 / 60 / 61 \better &
56 / 56 / 51 \muchworse &
55 / 55 / 58 \muchworse \\
\enskip instances with \example{the} \scriptsize{($=$1)} &
53 / 58 / 61 \inconsistent &
55 / 59 / 58 \inconsistent &
59 / 59 / 55 \muchworse &
63 / 63 / 63 \worse \\
\enskip instances with \example{a} \scriptsize{($\geq$1)} &
34 / 35 / 36 \muchworse &
34 / 36 / 37 \muchworse &
49 / 50 / 51 \muchworse &
48 / 52 / 50 \muchworse \\
\enskip instances with \example{two} \scriptsize{($\geq$2)} &
53 / 48 / 48 \worse &
50 / 50 / 49 \worse &
70 / 69 / 62 \worse &
72 / 67 / 58 \worse \\
\enskip instances with \example{most} &
49 / 50 / 49 \worse &
48 / 48 / 49 \worse &
69 / 68 / 60 \worse &
60 / 52 / 51 \muchworse \\
\enskip instances with \example{all} &
52 / 54 / 50 \worse &
48 / 50 / 51 \worse &
47 / 52 / 51 \muchworse &
49 / 50 / 51 \muchworse \\
\hline
\end{tabular}
\end{center}
\vspace{-0.2cm}
\caption{\label{figure:results}
Accuracy in \% (train/validation/test) of four selected models on our datasets, with a detailed evaluation of their ability to correctly understand specific instance types. Cell color indicates whether the corresponding instances were relatively \colorbox{red!35}{harder} or \colorbox{green!35}{easier} in comparison to the overall accuracy on the dataset, or whether the tendency is \colorbox{blue!25}{inconsistent} across train/validation/test accuracies.}
\end{figure*}

Figure \ref{figure:results} reports the train\footnote{Note that training accuracy here represents an interesting measure on its own, since no exact same instance is ever seen twice.}/validation/test performance of four models. In addition to the overall accuracy, it contains a detailed analysis of the models' ability to handle certain instance types. The accuracies for these \emph{dataset partitions} were obtained by restricting the dataset generator to sample an evaluation set of only one instance type.

Due to space limitations, we do not report detailed numbers for the other models. Essentially, they all show the same (or worse) behavior as the \textit{LSTM-only} model, apart from \textit{CNN+GRU:Mult} which is similar to \textit{CNN+LSTM:Mult}.

A number of conclusions from these results:
\begin{itemize}
\setlength{\topsep}{0cm}
\setlength{\itemsep}{0cm}
\setlength{\parsep}{0cm}
\item The consistently low performance (best: 60\%) indicates that all models essentially fail to learn spatial relations, in line with the findings of \newcite{Johnson2017}.\footnote{Contrary to what they report, our instances almost always require relational spatial reasoning.}
\item The numbers for the \textit{HCA} models on the \textsc{Quantification} dataset indicate that quantifiers are not fully learned. They may be approximated by a rough number/existence/majority estimate -- something we plan to investigate further.
\item Unsurprisingly, \textit{LSTM-only}, \textit{CNN-only} and also \textit{CNN+BoW:Mult} are not able to learn actual multimodal understanding, in contrast to their good performance on real-world data \cite{Jabri2016}. In our data, the failure in learning clearly shows in the tendency of these models to fall back to \textit{always-correct} or \textit{always-incorrect} predictions.
\item Although sometimes lower than training accuracy, the above-chance-level validation/test accuracy indicate that in some cases the models are able to generalize (to some degree).
\item Object recognition itself is not an issue -- the \textit{CNN-only} model trained for shape-color classification obtains $\sim$98\% accuracy.
\end{itemize}

There are many more interesting aspects that could be discussed, including learning curves, transfer learning and so on. However, the main point here is that a detailed investigation and error analysis like the one in figure \ref{figure:results} would be very difficult, if not impossible, to conduct with real-world data. It consequently shows the potential of artificial data as basis for a \emph{complementary} evaluation methodology for multimodal language understanding systems.

\subsection{Future work}
The basic \shapeworld\ framework can be elaborated in many ways. We plan to add new datasets addressing other aspects of language, as well as integrating options to enhance the language generation module, with the aim of providing more varied and natural image descriptions. For instance, we expect to integrate a subsequent step applying paraphrase rules after caption generation -- \newcite{Copestake2016} describe how this can be implemented on the level of DMRS graphs.

\section{Conclusion}

We have presented a new evaluation methodology and framework, \shapeworld, for multimodal deep learning models, with a focus on formal-semantic style generalization capabilities. In this framework, artificial data is automatically generated according to predefined specifications. This controlled data generation makes it possible to introduce previously unseen instance configurations during evaluation, which consequently require the system to recombine learned concepts in novel ways, i.e., true generalization.

We evaluated various VQA models on four image caption agreement datasets, where the system has to decide whether a statement applies to an image. We showed how the \shapeworld\ framework can be used to investigate in detail what these models learn with respect to multimodal language understanding. By exposing specific multimodal scenarios where current multimodal systems fail (e.g.\ spatial relations), and by providing \emph{a configurable, extensible testbed for systematic, detailed and comparable evaluation}, we hope to stimulate progress in the field of multimodal language understanding.

\bibliography{bibliography}

\begin{thebibliography}{}
\expandafter\ifx\csname natexlab\endcsname\relax\def\natexlab#1{#1}\fi

\bibitem[{Agrawal et~al.(2016)Agrawal, Batra, and Parikh}]{Agrawal2016}
Aishwarya Agrawal, Dhruv Batra, and Devi Parikh. 2016.
\newblock Analyzing the behavior of visual question answering models.
\newblock In {\em Proceedings of the 2016 Conference on Empirical Methods in
  Natural Language Processing\/}. Austin, Texas, EMNLP 2016, pages 1955--1960.

\bibitem[{Andreas et~al.(2016{\natexlab{a}})Andreas, Rohrbach, Darrell, and
  Klein}]{Andreas2016b}
Jacob Andreas, Marcus Rohrbach, Trevor Darrell, and Dan Klein.
  2016{\natexlab{a}}.
\newblock Learning to compose neural networks for question answering.
\newblock In {\em Proceedings of the Conference of the North American Chapter
  of the Association for Computational Linguistics\/}. NAACL 2016.

\bibitem[{Andreas et~al.(2016{\natexlab{b}})Andreas, Rohrbach, Darrell, and
  Klein}]{Andreas2016a}
Jacob Andreas, Marcus Rohrbach, Trevor Darrell, and Dan Klein.
  2016{\natexlab{b}}.
\newblock Neural module networks.
\newblock In {\em Proceedings of the IEEE Conference on Computer Vision and
  Pattern Recognition\/}. CVPR 2016.

\bibitem[{Antol et~al.(2015)Antol, Agrawal, Lu, Mitchell, Batra, Zitnick, and
  Parikh}]{Antol2015}
Stanislaw Antol, Aishwarya Agrawal, Jiasen Lu, Margaret Mitchell, Dhruv Batra,
  C.~Lawrence Zitnick, and Devi Parikh. 2015.
\newblock {VQA}: {V}isual question answering.
\newblock In {\em Proceedings of the IEEE International Conference on Computer
  Vision\/}. ICCV 2015.

\bibitem[{Arthur et~al.(2016)Arthur, Neubig, and Nakamura}]{Arthur2016}
Philip Arthur, Graham Neubig, and Satoshi Nakamura. 2016.
\newblock Incorporating discrete translation lexicons into neural machine
  translation.
\newblock In {\em Conference on Empirical Methods in Natural Language
  Processing (EMNLP)\/}. Austin, Texas, USA.

\bibitem[{Beattie et~al.(2016)Beattie, Leibo, Teplyashin, Ward, Wainwright,
  K{\"u}ttler, Lefrancq, Green, Vald{\'e}s, Sadik, Schrittwieser, Anderson,
  York, Cant, Cain, Bolton, Gaffney, King, Hassabis, Legg, and
  Petersen}]{DeepMindLab2016}
C.~Beattie, J.~Z. Leibo, D.~Teplyashin, T.~Ward, M.~Wainwright, H.~K{\"u}ttler,
  A.~Lefrancq, S.~Green, V.~Vald{\'e}s, A.~Sadik, J.~Schrittwieser,
  K.~Anderson, S.~York, M.~Cant, A.~Cain, A.~Bolton, S.~Gaffney, H.~King,
  D.~Hassabis, S.~Legg, and S.~Petersen. 2016.
\newblock Deep{M}ind {L}ab.
\newblock {\em ArXiv e-prints\/} .

\bibitem[{Bellemare et~al.(2013)Bellemare, Naddaf, Veness, and
  Bowling}]{Bellemare2013}
Marc~G. Bellemare, Yavar Naddaf, Joel Veness, and Michael Bowling. 2013.
\newblock The arcade learning environment: {A}n evaluation platform for general
  agents.
\newblock {\em Journal of Artificial Intelligence Research\/} 47(1):253--279.

\bibitem[{Bengio et~al.(1994)Bengio, Simard, and Frasconi}]{Bengio1994}
Yoshua Bengio, Patrice Simard, and Paolo Frasconi. 1994.
\newblock Learning long-term dependencies with gradient descent is difficult.
\newblock {\em Transactions on Neural Networks\/} 5(2):157--166.

\bibitem[{Bowman et~al.(2015)Bowman, Potts, and Manning}]{Bowman2015}
Samuel~R. Bowman, Christopher Potts, and Christopher~D. Manning. 2015.
\newblock Recursive neural networks can learn logical semantics.
\newblock In {\em Proceedings of the 3rd Workshop on Continuous Vector Space
  Models and their Compositionality\/}. Association for Computational
  Linguistics, Beijing.

\bibitem[{Brockman et~al.(2016)Brockman, Cheung, Pettersson, Schneider,
  Schulman, Tang, and Zaremba}]{OpenAI2016}
Greg Brockman, Vicki Cheung, Ludwig Pettersson, Jonas Schneider, John Schulman,
  Jie Tang, and Wojciech Zaremba. 2016.
\newblock Open{AI} {G}ym.

\bibitem[{Copestake et~al.(2016)Copestake, Emerson, Goodman, Horvat, Kuhnle,
  and Muszy{\'n}ska}]{Copestake2016}
Ann Copestake, Guy Emerson, Michael~W. Goodman, Matic Horvat, Alexander Kuhnle,
  and Ewa Muszy{\'n}ska. 2016.
\newblock Resources for building applications with {D}ependency {M}inimal
  {R}ecursion {S}emantics.
\newblock In {\em Proceedings of the 10th International Conference on Language
  Resources and Evaluation (LREC-16)\/}. European Language Resources
  Association (ELRA), Portoro\v{z}, Slovenia, pages 1240--1247.

\bibitem[{Copestake et~al.(2005)Copestake, Flickinger, Pollard, and
  Sag}]{Copestake2005}
Ann Copestake, Dan Flickinger, Carl Pollard, and Ivan~A. Sag. 2005.
\newblock Minimal {R}ecursion {S}emantics: {A}n introduction.
\newblock {\em Research on Language and Computation\/} 3(4):281--332.

\bibitem[{Flickinger(2000)}]{Flickinger2000}
Dan Flickinger. 2000.
\newblock On building a more efficient grammar by exploiting types.
\newblock {\em Natural Language Engineering\/} 6(1):15--28.

\bibitem[{Fukui et~al.(2016)Fukui, Park, Yang, Rohrbach, Darrell, and
  Rohrbach}]{Fukui2016}
Akira Fukui, Dong~Huk Park, Daylen Yang, Anna Rohrbach, Trevor Darrell, and
  Marcus Rohrbach. 2016.
\newblock Multimodal compact bilinear pooling for visual question answering and
  visual grounding.
\newblock In {\em Proceedings of the 2016 Conference on Empirical Methods in
  Natural Language Processing\/}. Austin, Texas, EMNLP 2016, pages 457--468.

\bibitem[{Gers and Schmidhuber(2001)}]{Gers2001}
Felix~A. Gers and J\"{u}rgen Schmidhuber. 2001.
\newblock {LSTM} recurrent networks learn simple context-free and
  context-sensitive languages.
\newblock {\em Transactions on Neural Networks\/} 12(6):1333--1340.

\bibitem[{Goyal et~al.(2016)Goyal, Khot, Summers-Stay, Batra, and
  Parikh}]{Goyal2016}
Yash Goyal, Tejas Khot, Douglas Summers-Stay, Dhruv Batra, and Devi Parikh.
  2016.
\newblock Making the {V} in {VQA} matter: {E}levating the role of image
  understanding in {V}isual {Q}uestion {A}nswering.
\newblock {\em CoRR\/} abs/1612.00837.

\bibitem[{He et~al.(2015)He, Zhang, Ren, and Sun}]{He2015}
Kaiming He, Xiangyu Zhang, Shaoqing Ren, and Jian Sun. 2015.
\newblock Delving deep into rectifiers: {S}urpassing human-level performance on
  {I}mage{N}et classification.
\newblock In {\em Proceedings of the IEEE International Conference on Computer
  Vision (ICCV)\/}. Santiago, Chile, pages 1026--1034.

\bibitem[{Hochreiter and Schmidhuber(1997)}]{Hochreiter1997}
Sepp Hochreiter and J\"{u}rgen Schmidhuber. 1997.
\newblock Long short-term memory.
\newblock {\em Neural Computation\/} 9(8):1735--1780.

\bibitem[{Hodosh and Hockenmaier(2016)}]{Hodosh2016}
Micah Hodosh and Julia Hockenmaier. 2016.
\newblock Focused evaluation for image description with binary forced-choice
  tasks.
\newblock In {\em Proceedings of the 5th Workshop on Vision and Language\/}.
  Berlin, Germany.

\bibitem[{Jabri et~al.(2016)Jabri, Joulin, and van~der Maaten}]{Jabri2016}
Allan Jabri, Armand Joulin, and Laurens van~der Maaten. 2016.
\newblock {\em Revisiting Visual Question Answering Baselines\/}, Springer
  International Publishing, Cham, pages 727--739.

\bibitem[{Johnson et~al.(2017)Johnson, Hariharan, van~der Maaten, Fei-Fei,
  Zitnick, and Girshick}]{Johnson2017}
Justin Johnson, Bharath Hariharan, Laurens van~der Maaten, Li~Fei-Fei,
  C.~Lawrence Zitnick, and Ross Girshick. 2017.
\newblock {CLEVR}: {A} diagnostic dataset for compositional language and
  elementary visual reasoning.
\newblock In {\em Computer Vision and Pattern Recognition (CVPR)\/}.

\bibitem[{Johnson et~al.(2016)Johnson, Hofmann, Hutton, and
  Bignell}]{Malmo2016}
Matthew Johnson, Katja Hofmann, Tim Hutton, and David Bignell. 2016.
\newblock The {M}almo platform for artificial intelligence experimentation.
\newblock In {\em Proceedings of the 25th International Joint Conference on
  Artificial Intelligence (IJCAI-16)\/}. AAAI Press, Palo Alto, California,
  USA, pages 4246--4247.

\bibitem[{Joulin and Mikolov(2015)}]{Joulin2015}
Armand Joulin and Tomas Mikolov. 2015.
\newblock Inferring algorithmic patterns with stack-augmented recurrent nets.
\newblock In {\em Advances in Neural Information Processing Systems 28\/},
  Curran Associates, Inc., pages 190--198.

\bibitem[{Karpathy and Li(2015)}]{Karpathy2015}
Andrej Karpathy and Fei{-}Fei Li. 2015.
\newblock Deep visual-semantic alignments for generating image descriptions.
\newblock In {\em Proceedings of the IEEE Conference on Computer Vision and
  Pattern Recognition\/}. CVPR 2015, pages 3128--3137.

\bibitem[{Kiela et~al.(2016)Kiela, Bulat, Vero, and Clark}]{Kiela2016}
Douwe Kiela, Luana Bulat, Anita~L. Vero, and Stephen Clark. 2016.
\newblock Virtual embodiment: {A} scalable long-term strategy for artificial
  intelligence research.
\newblock {\em CoRR\/} abs/1610.07432.

\bibitem[{Kingma and Ba(2014)}]{Kingma2014}
Diederik~P. Kingma and Jimmy Ba. 2014.
\newblock Adam: {A} method for stochastic optimization.
\newblock {\em CoRR\/} abs/1412.6980.

\bibitem[{Lu et~al.(2016)Lu, Yang, Batra, and Parikh}]{Lu2016}
Jiasen Lu, Jianwei Yang, Dhruv Batra, and Devi Parikh. 2016.
\newblock Hierarchical question-image co-attention for visual question
  answering.
\newblock In {\em Advances in Neural Information Processing Systems 29: Annual
  Conference on Neural Information Processing Systems 2016\/}. Barcelona,
  Spain, NIPS 2016, pages 289--297.

\bibitem[{Mikolov et~al.(2015)Mikolov, Joulin, and Baroni}]{Mikolov2015}
Tomas Mikolov, Armand Joulin, and Marco Baroni. 2015.
\newblock A roadmap towards machine intelligence.
\newblock {\em CoRR\/} abs/1511.08130.

\bibitem[{Montague(1970)}]{Montague1970}
Richard Montague. 1970.
\newblock English as a formal language.
\newblock In Bruno Visentini, editor, {\em Linguaggi nella societa e nella
  tecnica\/}, Edizioni di Communita, pages 188--221.

\bibitem[{Narasimhan et~al.(2015)Narasimhan, Kulkarni, and
  Barzilay}]{Narasimhan2015}
Karthik Narasimhan, Tejas Kulkarni, and Regina Barzilay. 2015.
\newblock Language understanding for text-based games using deep reinforcement
  learning.
\newblock In {\em Proceedings of the 2015 Conference on Empirical Methods in
  Natural Language Processing\/}. Association for Computational Linguistics,
  Lisbon, Portugal, pages 1--11.

\bibitem[{Nguyen et~al.(2015)Nguyen, Yosinski, and Clune}]{Nguyen2015}
Anh Nguyen, Jason Yosinski, and Jeff Clune. 2015.
\newblock Deep neural networks are easily fooled: {H}igh confidence predictions
  for unrecognizable images.
\newblock In {\em Proceedings of the IEEE Conference on Computer Vision and
  Pattern Recognition\/}. CVPR 2015.

\bibitem[{Pietroski et~al.(2009)Pietroski, Lidz, Hunter, and
  Halberda}]{Pietroski2009}
Paul Pietroski, Jeffrey Lidz, Tim Hunter, and Justin Halberda. 2009.
\newblock The meaning of 'most': {S}emantics, numerosity and psychology.
\newblock {\em Mind and Language\/} 24(5):554--585.

\bibitem[{Raposo et~al.(2017)Raposo, Santoro, Barrett, Pascanu, Lillicrap, and
  Battaglia}]{Raposo2017}
David Raposo, Adam Santoro, David Barrett, Razvan Pascanu, Timothy Lillicrap,
  and Peter~W. Battaglia. 2017.
\newblock Discovering objects and their relations from entangled scene
  representations.
\newblock {\em International Conference on Learning Representations 2017 (ICLR
  2017)\/} .

\bibitem[{Ren et~al.(2015)Ren, Kiros, and Zemel}]{Ren2015}
Mengye Ren, Ryan Kiros, and Richard~S. Zemel. 2015.
\newblock Image question answering: {A} visual semantic embedding model and a
  new dataset.
\newblock {\em CoRR\/} abs/1505.02074.

\bibitem[{Sorodoc et~al.(2016)Sorodoc, Lazaridou, Boleda, Herbelot, Pezzelle,
  and Bernardi}]{Sorodoc2016}
Ionut Sorodoc, Angeliki Lazaridou, Gemma Boleda, Aur\'{e}lie Herbelot, Sandro
  Pezzelle, and Raffaella Bernardi. 2016.
\newblock ``{L}ook, some green circles!'': {L}earning to quantify from images.
\newblock In {\em Proceedings of the 5th Workshop on Vision and Language\/}.
  Berlin, Germany.

\bibitem[{Sproat and Jaitly(2016)}]{Sproat2016}
Richard Sproat and Navdeep Jaitly. 2016.
\newblock {{RNN} Approaches to Text Normalization: {A} Challenge}.
\newblock {\em CoRR\/} abs/1611.00068.

\bibitem[{Sukhbaatar et~al.(2015)Sukhbaatar, Szlam, Synnaeve, Chintala, and
  Fergus}]{Sukhbaatar2015}
Sainbayar Sukhbaatar, Arthur Szlam, Gabriel Synnaeve, Soumith Chintala, and Rob
  Fergus. 2015.
\newblock Maze{B}ase: {A} sandbox for learning from games.
\newblock {\em CoRR\/} abs/1511.07401.

\bibitem[{Szegedy et~al.(2014)Szegedy, Zaremba, Sutskever, Bruna, Erhan,
  Goodfellow, and Fergus}]{Szegedy2014}
Christian Szegedy, Wojciech Zaremba, Ilya Sutskever, Joan Bruna, Dumitru Erhan,
  Ian~J. Goodfellow, and Rob Fergus. 2014.
\newblock Intriguing properties of neural networks.
\newblock {\em CoRR\/} abs/1312.6199.

\bibitem[{Vinyals et~al.(2015)Vinyals, Fortunato, and Jaitly}]{Vinyals2015}
Oriol Vinyals, Meire Fortunato, and Navdeep Jaitly. 2015.
\newblock Pointer networks.
\newblock In {\em Proceedings of the 28th International Conference on Neural
  Information Processing Systems\/}. MIT Press, Montreal, Canada, NIPS'15,
  pages 2692--2700.

\bibitem[{Weston et~al.(2015)Weston, Bordes, Chopra, and Mikolov}]{Weston2015}
Jason Weston, Antoine Bordes, Sumit Chopra, and Tomas Mikolov. 2015.
\newblock Towards {AI}-complete question answering: {A} set of prerequisite toy
  tasks.
\newblock {\em CoRR\/} abs/1502.05698.

\bibitem[{Zaremba and Sutskever(2014)}]{Zaremba2014}
Wojciech Zaremba and Ilya Sutskever. 2014.
\newblock Learning to execute.
\newblock {\em CoRR\/} abs/1410.4615.

\bibitem[{Zhang et~al.(2017)Zhang, Bengio, Hardt, Recht, and
  Vinyals}]{Zhang2017}
Chiyuan Zhang, Samy Bengio, Moritz Hardt, Benjamin Recht, and Oriol Vinyals.
  2017.
\newblock Understanding deep learning requires rethinking generalization.
\newblock {\em International Conference on Learning Representations 2017 (ICLR
  2017)\/} .

\bibitem[{Zhang et~al.(2016)Zhang, Goyal, Summers{-}Stay, Batra, and
  Parikh}]{Zhang2016}
Peng Zhang, Yash Goyal, Douglas Summers{-}Stay, Dhruv Batra, and Devi Parikh.
  2016.
\newblock {Y}in and {Y}ang: {B}alancing and answering binary visual questions.
\newblock In {\em Conference on Computer Vision and Pattern Recognition
  (CVPR)\/}.

\bibitem[{Zitnick and Parikh(2013)}]{Zitnick2013b}
C.~Lawrence Zitnick and Devi Parikh. 2013.
\newblock Bringing semantics into focus using visual abstraction.
\newblock In {\em Proceedings of the IEEE Conference on Computer Vision and
  Pattern Recognition\/}. CVPR 2013, pages 3009--3016.

\bibitem[{Zitnick et~al.(2016)Zitnick, Vedantam, and Parikh}]{Zitnick2016}
C.~Lawrence Zitnick, Ramakrishna Vedantam, and Devi Parikh. 2016.
\newblock Adopting abstract images for semantic scene understanding.
\newblock {\em IEEE Transactions on Pattern Analysis and Machine
  Intelligence\/} 38(4):627--638.

\end{thebibliography}
\bibliographystyle{emnlp_natbib}
\end{document}